\documentclass[runningheads]{llncs}
\usepackage[T1]{fontenc}
\usepackage{graphicx,verbatim}
\usepackage{amsmath, amssymb}
\usepackage{booktabs}
\usepackage{color}
\usepackage{hyperref}

\begin{document}
\title{Learning Cross-Joint Attention for Generalizable Video-Based Seizure Detection\thanks{Official implementation of the proposed method is available on \href{https://github.com/OmarSZamzam/joint-attention-seizure-detection}{GitHub}.}}
\titlerunning{Learning Body Dynamics for Seizure Detection}
\begin{comment}  %% Removed for anonymized MICCAI submission
\author{First Author\inst{1}\orcidID{0000-1111-2222-3333} \and
Second Author\inst{2,3}\orcidID{1111-2222-3333-4444} \and
Third Author\inst{3}\orcidID{2222--3333-4444-5555}}
%
\authorrunning{F. Author et al.}
\institute{Princeton University, Princeton NJ 08544, USA \and
Springer Heidelberg, Tiergartenstr. 17, 69121 Heidelberg, Germany
\email{lncs@springer.com}\\
\url{http://www.springer.com/gp/computer-science/lncs} \and
ABC Institute, Rupert-Karls-University Heidelberg, Heidelberg, Germany\\
\email{\{abc,lncs\}@uni-heidelberg.de}}

\end{comment}

\author{Omar Zamzam \and
Takfarinas Medani \and
Chinmay Chinara \and
Richard M. Leahy}

\authorrunning{O. Zamzam et al.}
\institute{Ming Hsieh Department of Electrical and Computer Engineering \\
University of Southern California \\
Los Angeles, CA 90089 \\
\email{\{zamzam, medani, chinara, leahy\}@usc.edu}}
  
\maketitle              % typeset the header of the contribution
\begin{abstract}
Automated seizure detection from long-term clinical videos can substantially reduce manual review time and enable real-time monitoring. However, existing video-based methods often struggle to generalize to unseen subjects due to background bias and reliance on subject-specific appearance cues. We propose a joint-centric attention model that focuses exclusively on body dynamics to improve cross-subject generalization. For each video segment, body joints are detected and joint-centered clips are extracted, suppressing background context. These joint-centered clips are tokenized using a Video Vision Transformer (ViViT), and cross-joint attention is learned to model spatial and temporal interactions between body parts, capturing coordinated movement patterns characteristic of seizure semiology. Extensive cross-subject experiments show that the proposed method consistently outperforms state-of-the-art CNN-, graph-, and transformer-based approaches on unseen subjects.

% Authors must provide keywords and are not allowed to remove this Keyword section.

\end{abstract}

\section{Introduction}

Continuous video monitoring is a central component of Epilepsy Monitoring Units (EMUs), where patients are recorded over extended periods ranging from several days to multiple weeks \cite{basnyat2022clinical,cascino2002video,kobulashvili2018diagnostic}. These long-term recordings support critical clinical tasks such as seizure semiology analysis and real-time monitoring for timely seizure detection and intervention. In both settings, automated video-based seizure detection systems have the potential to substantially reduce clinical workload, and improve patient safety.

Despite this clinical importance, automated video-based seizure detection remains challenging. Seizure-related motor manifestations exhibit significant variability across patients, seizure types, and anatomical origins \cite{fisher2017operational,zuberi2022ilae,beniczky2022seizure}. While human observers can reliably identify seizures by focusing on characteristic involuntary movements—such as rhythmic jerking, tonic stiffening, or coordinated multi-limb contractions—many deep learning-based video classification methods struggle to generalize across subjects. This limitation is often attributed to reliance on spurious visual cues, such us background appearance, camera viewpoint, and subject-specific morphology, rather than on seizure-related motion patterns.

In this work, we argue that a generalizable video-based seizure detector should explicitly model \emph{motor dynamics} while treating non-motor visual information as background to be ignored. Seizure-related motor activity arises from involuntary muscle activations that may involve individual joints or coordinated motion across multiple body parts, with relationships that vary across seizure types, severity levels, and subjects \cite{tufenkjian2012seizure,so2006value,rossetti2010seizure}. Consequently, enforcing static correlations between body parts can introduce bias and hinder generalization.

Motivated by this observation, we propose a joint-centric video-based seizure detection framework that decomposes each video segment into multiple joint-centered sub-videos. Each joint-specific video is processed independently using a pretrained \emph{video} vision transformer to obtain a compact latent representation capturing the temporal dynamics of that joint. These joint-level representations, interpreted as motion tokens, are then integrated using a learned cross-joint attention mechanism that dynamically captures coordinated motor patterns. Unlike graph-based approaches with fixed adjacency structures, our attention-based formulation allows joint relationships to adapt based on observed motion, enabling robustness to subject identity, body morphology, and recording environment.

We evaluate the proposed method on a public long-term seizure video dataset introduced in \cite{xu2024vsvig} using a strict subject-wise train–test split. Experimental results demonstrate that the proposed joint-centric framework substantially outperforms state-of-the-art video-based seizure detection methods across multiple evaluation metrics, achieving strong generalization to unseen subjects.

\textbf{Contributions.} The main contributions of this work are:
\begin{itemize}
    \item We leverage pretrained video vision transformer models to encode joint-level motion into compact and informative latent tokens.
    \item We propose a learned cross-joint attention mechanism that captures dynamic and subject-adaptive motor coordination patterns.
    \item We demonstrate improved performance compared to state-of-the-art methods on seizure detection from long-term clinical videos.
\end{itemize}

\section{Related Work}

Automated seizure detection has been studied using a variety of automated video classification approaches \cite{jory2016safe} \cite{van2016non} \cite{rai2024automated}. More recent methods commonly rely on convolutional or transformer-based deep learning methods operating on full video frames to capture spatiotemporal patterns associated with seizure activity \cite{karacsony2022novel} \cite{boyne2025video} \cite{aslani2026video} \cite{ogura2015neural}. These models often struggle to generalize to unseen subjects due to their sensitivity to background appearance, recording environment, and subject-specific visual cues.

To mitigate background bias, skeleton- and pose-based methods have been proposed that explicitly model human motion \cite{huang2023posture} \cite{karayiannis2006automated} \cite{karacsony2025exploring}. A representative example is VSViG \cite{xu2024vsvig}, which employs a spatiotemporal graph representation of body joints for real-time seizure detection and demonstrates improved cross-subject performance compared to full-frame video models. However, many skeleton-based approaches rely on predefined graph structures with fixed joint connectivity, which may not adequately capture the diverse and dynamic coordination patterns observed across different seizure types and subjects \cite{fisher2017operational} \cite{moshe2015epilepsy} \cite{thijs2019epilepsy}.

Recent advances in large-scale video vision foundation models, such as Hiera \cite{ryali2023hiera}, have enabled learning rich spatiotemporal representations transferable to downstream tasks. In the context of seizure detection, foundation models have been applied to clinical and smartphone-recorded videos, showing promising results \cite{miron2025detection}. Nevertheless, existing approaches typically apply these models to entire video frames, leaving them susceptible to the same non-motor visual biases that limit generalization.

In contrast to prior work, our method models seizure activity as a coordinated motor phenomenon by decomposing videos into joint-centric representations and learning adaptive relationships between body parts. By combining pretrained video vision transformer motion encoding with a learned cross-joint attention mechanism, the proposed approach captures seizure-relevant dynamics while discarding non-informative visual context, enabling improved generalization across subjects.

\section{Method}

\begin{figure}[t]
    \centering
    \includegraphics[width=\textwidth]{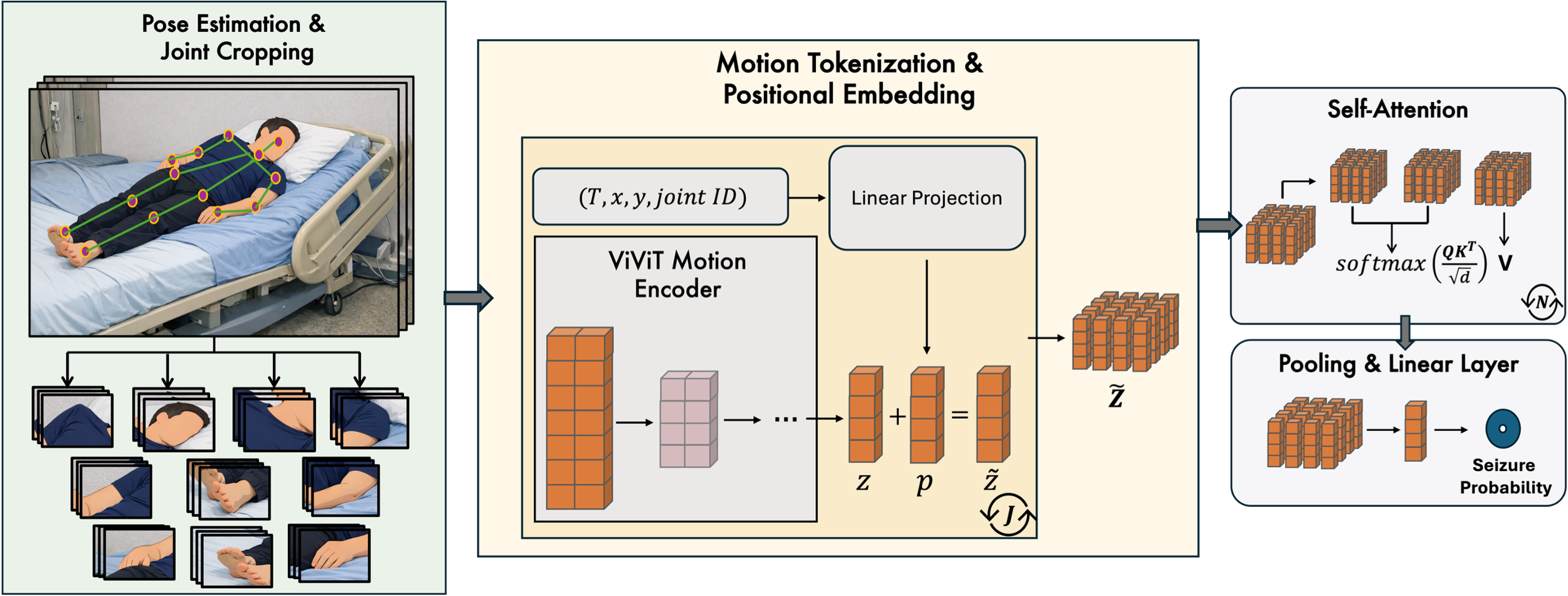}
    \caption{
    Overview of the proposed joint-centric video-based seizure detection framework.
    A clinical video segment is first processed using pose estimation to localize major body joints, and joint-centric sub-videos are extracted for each body part.
    Each joint video is encoded independently using a shared pretrained Video Vision Transformer (ViViT) to obtain joint-level motion tokens.
    Positional information, including joint location and joint identity, is projected and added to the motion tokens.
    A multi-head self-attention module models adaptive inter-joint relationships, and the resulting representations are pooled and passed through a linear classification layer to predict seizure presence.
    }
    \label{fig:seizure_detection}
\end{figure}

Given a video segment $\mathbf{x} \in \mathbb{R}^{T \times H \times W \times 3}$ corresponding to a temporal window of T = 5 seconds,  $H=1080$, $W=1920$, and the three channels correspond to RGB, the proposed method models seizure activity by explicitly isolating and correlating motor dynamics across major body joints. Videos are originally recorded at 30 frames per second (fps) and are temporally downsampled to 6 fps, resulting in $T=30$ frames per segment. The central idea is to represent seizure-related motion as a set of joint-level temporal tokens and to learn their relationships through an adaptive attention mechanism, while discarding non-motor visual information such as background appearance and subject-specific context.

\subsection{Joint-Centric Video Representation}

We first use a pretrained pose estimation model (OpenPose \cite{cao2019openpose}) on each video frame to localize a set of $J=14$ clinically relevant body joints, including the head, neck, shoulders, elbows, wrists, hips, knees, and ankles. For each joint $j \in \{1,\dots,J\}$, we extract a joint-centric video $\mathbf{x}_j \in \mathbb{R}^{T \times h \times w \times 3}$ by cropping a fixed-size ($h=w=120$) spatial window centered at the joint location in every frame. This produces a set of temporally aligned joint videos $\{\mathbf{x}_1, \dots, \mathbf{x}_J\}$, each capturing the motion of a specific body part while minimizing background and contextual visual cues as illustrated in the left-most panel in Fig. \ref{fig:seizure_detection}.

Each joint-centric video is processed independently using  the encoder of a pretrained video vision transformer, Video Vision Transformer (ViViT) \cite{arnab2021vivit}, which serves as a motion encoder. All joints share the same ViViT weights. Given a joint video $\mathbf{x}_j$, the encoder $E$ produces a latent representation
\begin{equation}
    \mathbf{z}_j = E_{\text{ViViT}}(\mathbf{x}_j) \in \mathbb{R}^{d},
\end{equation}
where $\mathbf{z}_j$ summarizes the temporal dynamics of joint $j$ over the duration of the segment. These latent vectors can be interpreted as joint-level motion tokens, each encoding the behavior of a specific body part.

In addition to motion features, we incorporate explicit positional information for each joint to provide global spatial context. For each joint-centric video, we construct a positional encoding tensor of shape $(J, T, 3)$ that contains the spatial coordinates $(x,y)$ of the joint location in the original uncropped video, along with a joint identity indicator. The spatial coordinates convey the relative location of the joint within the full body configuration, while the joint identity indicates the anatomical role of each token. This positional information is passed through a learnable linear projection that maps it to the same embedding dimension $d$. The resulting positional embeddings are added to the joint-level motion tokens to form position-augmented tokens
\begin{equation}
    \tilde{\mathbf{z}}_j = \mathbf{z}_j + \mathbf{p}_j, \quad j = 1,\dots,J,
\end{equation}
where $\mathbf{p}_j \in \mathbb{R}^{d}$ denotes the projected positional embedding for joint $j$.

\subsection{Cross-Joint Attention and Classification}

To model seizure activity as a coordinated motor phenomenon, the set of position-augmented joint tokens $\tilde{\mathbf{Z}} = \{\tilde{\mathbf{z}}_1, \dots, \tilde{\mathbf{z}}_J\}$ is processed using a multi-head self-attention mechanism. Unlike graph-based approaches with fixed joint connectivity, attention enables the model to learn adaptive and context-dependent relationships between joints. Formally, the attention operation is defined as
\begin{equation}
    \text{Attn}(\tilde{\mathbf{Z}}) = \text{softmax}\left(\frac{\mathbf{Q}\mathbf{K}^\top}{\sqrt{d}}\right)\mathbf{V},
\end{equation}
where queries $\mathbf{Q}$, keys $\mathbf{K}$, and values $\mathbf{V}$ are linear projections of the position-augmented joint tokens, and $d$ is the token embedding dimension used for scaling. This formulation enables the model to emphasize joint interactions that are most informative for seizure detection, allowing correlations between body parts to vary across subjects and seizure types.

The output of the attention module is pooled to form a global representation $\mathbf{u}$ of the video segment. A linear classification head is then applied to produce a scalar logit
\begin{equation}
    \ell = \mathbf{w}^\top \mathbf{u} + b,
\end{equation}
which is used to predict seizure presence. The model is trained using binary cross-entropy loss on the logits.

\subsection{Parameter-Efficient Fine-Tuning}

To adapt the pretrained ViViT encoder to the seizure detection task while avoiding overfitting on limited clinical data, we consider two training strategies. In the first, all ViViT parameters are frozen and only the attention module and classification head are trained. In the second, we apply Low-Rank Adaptation (LoRA) \cite{hu2022lora} to the ViViT encoder. Specifically, LoRA modules are injected into the query, key, value, and feed-forward linear layers of the last two transformer blocks of ViViT, which are closest to the token output. This restricts adaptation to higher-level motion representations while keeping the majority of pretrained parameters fixed. LoRA introduces trainable low-rank matrices with rank $r=8$, scaling factor $\alpha=16$, and dropout $0.05$, enabling parameter-efficient fine-tuning with minimal computational overhead.

By decomposing videos into joint-centric motion representations, augmenting them with explicit positional information, and learning adaptive inter-joint relationships, the proposed method focuses explicitly on seizure-related motor dynamics while remaining invariant to background appearance, body morphology, and recording environment. This design enables improved generalization to unseen subjects and diverse seizure manifestations.

\section{Experimental Setup}

\paragraph{Dataset.}
We evaluate the proposed method on a public long-term seizure video dataset originally introduced in VSViG \cite{xu2024vsvig}. The dataset consists of 33 seizure videos recorded in EMUs, with video durations ranging from approximately 30 seconds to over one hour. The videos are collected from 14 subjects, including 7 subjects with focal (partial) seizures and 7 subjects with focal seizures that generalize to tonic-clonic seizures. Each video is annotated with EEG onset time as well as clinical onset time which corresponds to the moment when clear pathological motor behavior becomes observable. In the provided dataset, the faces of all subjects are blurred to ensure subject de-identification.

\paragraph{Segment construction and labeling.}
Each long video is divided into fixed-length segments of 5 seconds. Videos are recorded at 30 fps and temporally downsampled to 6 fps, resulting in 30 frames per segment. Segments ending before the EEG onset time are labeled as \emph{interictal}. Segments starting within a 40-second window following the clinical onset time are labeled as \emph{ictal}.

\paragraph{Train-validation-test split.}
We follow a subject-wise evaluation protocol to assess generalization to unseen subjects. Subjects are split into disjoint sets, with 9 subjects used for training and validation and 5 held-out subjects used exclusively for testing. This results in 1,770 training segments (1,062 interictal, 708 ictal) and 952 test segments (565 interictal, 387 ictal). Among the test subjects, 2 exhibit focal seizures and 3 exhibit focal-to-generalized tonic-clonic seizures.

\paragraph{Baselines.}
We compare the proposed method against state-of-the-art video-based seizure detection approaches, including VSViG \cite{xu2024vsvig}, a skeleton-based spatiotemporal graph model designed for real-time seizure detection, Fine-tuned Hiera a foundation-model-based video classifier proposed for epileptic spasm detection from video \cite{miron2025detection}, 3D temporal CNN inspired by \cite{yang2021video}, and 3D temporal CNN applied on optical flows extracted from the raw videos which is inspired by \cite{boyne2025video}, \cite{geertsema2018automated}, and \cite{mehta2023privacy}. In addition, we report results for a random classifier and a naive classifier that predicts all segments as one class.

\paragraph{Evaluation metrics.}
Performance is evaluated using accuracy, area under the receiver operating characteristic curve (AUROC), area under the precision-recall curve (AUPRC), F1 score, precision, and recall. Given the class imbalance inherent in long-term seizure monitoring data, AUROC and AUPRC are emphasized as primary evaluation metrics.

\section{Results}

\begin{table}[t]
\centering
\footnotesize
\setlength{\tabcolsep}{6pt}
\caption{Performance comparison on the held-out test subjects reported as mean $\pm$ standard deviation across 5 runs.}
\label{tab:results}
\begin{tabular}{lccc}
% \toprule
% \multicolumn{4}{c}{\textbf{Overall Detection Performance}} \\
\midrule
Method & Accuracy & AUROC & AUPRC \\
\midrule
Random Chance 
& $0.510\pm0.016$ & $0.514\pm0.018$ & $0.415\pm0.011$ \\
One-Class Classifier
& $0.593\pm0.000$ & $0.500\pm0.000$ & $0.407\pm0.000$ \\
VSViG
& $0.678\pm0.018$ & $0.732\pm0.019$ & $0.691\pm0.019$ \\
3D-CNN
& $0.790\pm0.028$ & $0.772\pm0.037$ & $0.735\pm0.075$ \\
O.F. 3D-CNN
& $0.736\pm0.068$ & $0.758\pm0.050$ & $0.714\pm0.051$ \\
Hiera + LORA
& $0.814\pm0.016$ & $0.868\pm0.009$ & $0.743\pm0.013$ \\
Proposed
& $0.866\pm0.017$ & $0.921\pm0.012$ & $0.870\pm0.018$ \\
Proposed + LORA
& $\mathbf{0.889\pm0.010}$ & $\mathbf{0.923\pm0.012}$ & $\mathbf{0.881\pm0.022}$ \\
\midrule
\multicolumn{4}{c}{} \\
\midrule
Method & F1 & Precision & Recall \\
\midrule
Random Chance 
& $0.452\pm0.021$ & $0.414\pm0.018$ & $0.498\pm0.027$ \\
One-Class Classifier
& $0.578\pm0.000$ & $0.407\pm0.000$ & $1.000\pm0.000$ \\
VSViG
& $0.665\pm0.021$ & $0.608\pm0.045$ & $0.737\pm0.015$ \\
3D-CNN
& $0.721\pm0.013$ & $0.809\pm0.119$ & $0.666\pm0.078$ \\
O.F. 3D-CNN
& $0.676\pm0.056$ & $0.691\pm0.101$ & $0.685\pm0.060$ \\
Hiera + LORA
& $0.730\pm0.015$ & $0.669\pm0.041$ & $0.786\pm0.058$ \\
Proposed
& $0.841\pm0.019$ & $0.814\pm0.039$ & $0.872\pm0.054$ \\
Proposed + LORA
& $\mathbf{0.864\pm0.012}$ & $\mathbf{0.854\pm0.006}$ & $\mathbf{0.875\pm0.024}$ \\
\bottomrule
\end{tabular}
\end{table}

Table~\ref{tab:results} reports the performance of all evaluated methods on the held-out test subjects, averaged over five runs. The proposed joint-centric approach achieves the highest performance across all reported metrics, both with frozen ViViT features and with LoRA-based fine-tuning.

Using frozen video transformer features, the proposed method substantially outperforms skeleton-based and full-frame video baselines in terms of accuracy, AUROC, AUPRC, F1 score, precision, and recall. Applying parameter-efficient fine-tuning via LoRA further improves performance across all metrics, yielding the best overall results.

Across all baselines, graph-based and convolutional models exhibit lower precision and F1 scores compared to the proposed approach, while maintaining comparable recall. These results highlight consistent performance gains of the proposed method across both threshold-free and threshold-dependent evaluation metrics.

\begin{figure}[t]
    \centering
    \includegraphics[width=\textwidth]{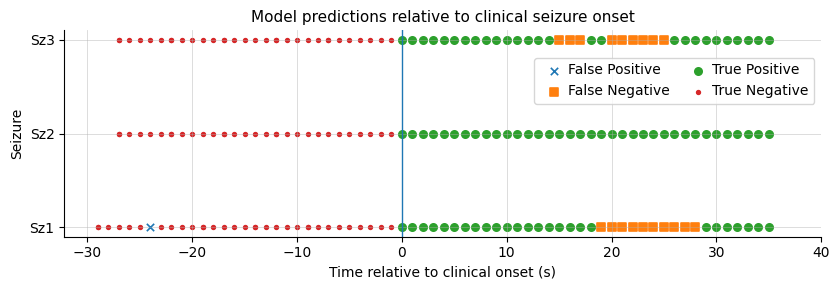}
    \caption{Temporal distribution of model predictions relative to clinical onset for one subject. Predictions are produced at 1\,s intervals.}
    \label{fig:error_timing}
\end{figure}

Fig.~\ref{fig:error_timing} shows the temporal distribution of model predictions aligned to clinical onset ($t{=}0$) for the test subject with the largest number of seizures (3 events). The frames corresponding to consecutive false negatives in Seizure 3 showed partial occlusion of the patient feature-points by intervening medical staff.

\section{Discussion}

The results demonstrate that modeling seizure activity as a coordinated motor movements improves generalization to unseen subjects. By decomposing videos into joint-centric representations and learning adaptive inter-joint relationships through self-attention, the proposed method outperforms both full-frame video classifiers and skeleton-based graph models. This supports the premise that explicitly focusing on seizure-related motor dynamics while suppressing background and subject-specific visual cues leads to more robust detection.

Compared to methods with fixed skeletal connectivity, the proposed attention-based formulation enables flexible modeling of inter-joint dependencies that can adapt to heterogeneous seizure manifestations across subjects and seizure types, while capturing fewer subject-specific characteristics. The strong performance achieved using frozen video transformer features highlights the effectiveness of pretrained video representations and their sufficient descriptive power, while parameter-efficient fine-tuning via LoRA provides additional gains with minimal increase in trainable parameters.

It is worth noting that the proposed approach is mainly designed to model motor manifestations through coordinated joint motion and has not been evaluated on non-motor seizures characterized by subtle or absent movement patterns. Extending the framework to such seizure types would require evaluation on datasets that preserve visual cues such as facial features, to enable the incorporation of information related to gaze or fine-grained facial motion.

\section{Conclusion}

We presented a joint-centric video-based seizure detection framework that models seizure activity through adaptive inter-joint relationships. By extracting joint-centric video representations, encoding motion using a pretrained video vision transformer, and applying self-attention to capture coordinated body dynamics, the proposed method outperforms state-of-the-art performance on a public long-term seizure video dataset under subject-wise evaluation.

The results demonstrate that focusing on seizure-related motor dynamics while discarding non-informative visual context leads to improved generalization across subjects. These findings suggest that joint-centric attention-based modeling is a promising direction for robust video-based seizure detection and has the potential to support both clinical review and automated monitoring systems in epilepsy care.

\bibliographystyle{splncs04}
\bibliography{mybibliography}

\end{document}